\documentclass[lettersize,journal]{IEEEtran}
\usepackage{amsmath,amsfonts}
\usepackage[ruled,vlined]{algorithm2e}
\usepackage{array}
\usepackage{subcaption}

\usepackage{outlines}
\usepackage{caption}
\usepackage{multirow}
\usepackage{hyperref}
\usepackage{comment}
\usepackage{enumerate}
\usepackage{booktabs}

\usepackage{textcomp}
\usepackage{stfloats}
\usepackage{url}
\usepackage{verbatim}
\usepackage{graphicx}
\usepackage{cite}
\begin{document}

\title{Spatial Clustering Approach for Vessel Path Identification}

\author{Mohamed Abuella,~\IEEEmembership{} M. Amine Atoui,~\IEEEmembership{} Slawomir Nowaczyk,~\IEEEmembership{} 
Simon Johansson,~\IEEEmembership{}and Ethan Faghani.~\IEEEmembership{}

\thanks{Manuscript received November 1, 2023; revised November 16, 2023.}

\thanks{This work was supported by the Sweden's innovation agency (Vinnova).(\textit{Corresponding author: Mohamed Abuella}.)\\
Mohamed Abuella, M. Amine Atoui, and Slawomir Nowaczy are with Center for Applied Intelligent Systems Research (CAISR), Halmstad University, Kristian IV:s väg 80523, Halmstad, 30118 Sweden (e-mail: mohamed.abuella@hh.se;
amine.atoui; Slawomir.Nowaczyk@hh.se).
Simon Johansson and Ethan Faghani are with CetaSol AB, Gothenburg, 41251, Sweden (e-mail: simon.johansson and ethan.faghani\}@cetasol.com)}}

\markboth{Journal of \LaTeX\ Class Files,~Vol.~14, No.~8, August~2021}%
{Shell \MakeLowercase{\textit{et al.}}: A Sample Article Using IEEEtran.cls for IEEE Journals}


\maketitle
\begin{abstract}
This paper addresses the challenge of identifying the paths for vessels with operating routes of repetitive paths, partially repetitive paths, and new paths. We propose a spatial clustering approach for labeling the vessel paths by using only position information. 
We develop a path clustering framework employing two methods: a distance-based path modeling and a likelihood estimation method. The former enhances the accuracy of path clustering through the integration of unsupervised machine learning techniques, while the latter focuses on likelihood-based path modeling and introduces segmentation for a more detailed analysis.
The result findings highlight the superior performance and efficiency of the developed approach, as both methods for clustering vessel paths into five classes achieve a perfect F1-score.
The approach aims to offer valuable insights for route planning, ultimately contributing to improving safety and efficiency in maritime transportation.
\end{abstract}


\maketitle

\section{Introduction}\label{Sec:Intro}
\subsection{Background and Motivation}\label{Sec:Back}
\IEEEPARstart{T}{he} maritime transportation is crucial for global trade, generating extensive vessel trajectory data that reveals intricate spatial and temporal navigation patterns. Understanding these patterns is vital for effective maritime traffic surveillance and management~\cite{tu2017exploiting}.

A \textit{trajectory} refers to a sequence of consecutive geographical points, each representing a specific location at a given timestamp~\cite{zheng2015trajectory}. Trajectories, observed in various scenarios such as pedestrian movements, vehicular routes, and natural events like wildlife migrations or hurricanes, involve time-evolving position data. Trajectory mining aims to uncover significant patterns within datasets, enabling tasks like path classification, anomaly detection for accidents or traffic congestions, surveillance for suspicious activities, and prediction of vessel trajectories in different landscapes~\cite{lee2007trajectory}.

It is crucial to distinguish between trajectory and path when studying movement data. The term trajectory typically denotes the movement of an object over time, while \textit{path} refers to the specific route taken by the object. For instance, if an object travels from origin to destination, its trajectory is the sequence of geographical points it passes through, while its path is the specific route taken, such as a sidewalk, street, highway, lane, railway, or waterway.

Additionally, the term \textit{route} is synonymous with path or trajectory as long as it shares the same origin and destination. Conversely, any difference in either the origin or destination represents another route.

Path clustering, a versatile technique, involves grouping paths into clusters based on their similarity.
This approach has found its niche in a wide range of practical applications. In the realm of navigation, path identification empowers systems to generate clear and detailed instructions for users seeking their way. Traffic analysis benefits from path clustering as it facilitates the identification of diverse traffic patterns, such as the smooth flow of traffic on highways and the congestion often encountered on city streets. Path identification proves equally valuable in route planning, enabling the optimization of routes for transportation systems, including public and maritime transportation services~\cite{zheng2011computing}.

In the scope of the maritime industry, path identification from Automatic Identification System (AIS) data is a challenging task due to the high spatial freedom and, especially in coastal areas, the high frequency of ship's navigation maneuvers. Thus, it is imperative to develop a path identification tool that integrates with route planning systems for improving maritime safety and optimizing vessel routing. As data-driven approaches from AIS data continue to grow and evolve, path clustering will undoubtedly play an increasingly important role in understanding vessel behavior and supporting decision-making in maritime transportation~\cite{yan2021emerging}.

\subsection{Aims and Contributions}\label{Sec:aim_cont}

This paper aims to address the challenge of identifying vessel paths in scenarios characterized by repetitive, semi-repetitive, and novel operations. 
In general, the aims and contributions of the proposed approach in this paper can be outlined as follows:
\begin{itemize}
    \item The proposed clustering approach of vessel paths requires only position information, specifically longitude and latitude.
    \item The clustering approach has a proven added value for clustering challenging unseen or unknown paths.
    \item The approach is robust and interpretable by applying a similarity measure that reduces the influence of noise or outliers and offers a clear interpretation of path clustering.
    \item The approach has a customizable parameter to determine the number of path classes, thereby enhancing the flexibility and adaptability of the framework and allowing users to tailor it to their specific needs.
    \item The approach also includes a method to study and analyze the patterns within specific segments of a path.
    \item It is a data-driven solution that can be used as a valuable asset for informed decision-making in route planning and optimization, traffic management, and resource allocation.
\end{itemize}

The rest of this paper is organized as follows. Section~\ref{Subsec:related_work} reviews the related work on vessel path identification. Section~\ref{Sec:Methodology} describes our proposed spatial clustering approach in detail. The real-world vessel data of our case study is described in Section~\ref{Sec:Case_study}. Section~\ref{Sec:Results} presents the results of our experimental evaluation. Section~\ref{Sec:Conclusion} concludes the paper and discusses future work.

\section{Related Work}\label{Subsec:related_work}
Path clustering can be done using a variety of different methods~\cite{yuan2017review}. We will explore the related works of these various methods.

Clustering is gaining popularity for route extraction. Machine learning (ML) has recently been applied extensively for vessel path identification by learning patterns from historical data. 
The study presented in~\cite{pallotta2013vessel} introduced a framework, Traffic Route Extraction and Anomaly Detection (TREAD), which utilizes unsupervised learning for maritime route extraction. The primary emphasis is on anomaly detection and route prediction, highlighting the crucial role of AIS data in enhancing maritime situational awareness. The work specifically addresses challenges related to intermittency and persistence in AIS data.
Another method of route extraction was proposed in~\cite{Yan2020}, transforming ship trajectories into ship trip semantic objects (STSO) and utilizing graph theory for route extraction. The method proves robustness in extracting traffic routes for merchant ships but may have limitations for vessels with frequent navigation behavior changes, such as fishing vessels.
The approach in~\cite{zhao2019novel}, on the other hand, adopts a dynamic time warping (DTW) distance as a similarity measure and considers vessel course changes to analyze its trajectories. Experiments demonstrated its high accuracy in distinguishing and detecting similar vessel trajectories, outperforming existing methods in accuracy and cluster degree evaluation.
Moreover,~\cite{de2012machine} presents a machine learning framework for maritime vessel trajectory analysis, incorporating clustering, classification, and outlier detection. It employs piecewise linear segmentation for compression and alignment kernels to integrate geographical domain knowledge, enhancing task performance. Results show reduced computation time without compromising accuracy.

Capobianco et al.\cite{capobianco2021deep} proposed a deep learning approach using recurrent neural networks, employing a Bidirectional Long Short-Term Memory (BiLSTM) layer as an encoder and a Unidirectional Long Short-Term Memory (LSTM) layer as a decoder, for vessel trajectory prediction. Their model outperforms baseline approaches, showcasing the effectiveness of sequence-to-sequence neural networks.
In their study, Li et al.~\cite{li2023incorporation} present an AIS data-based machine learning method for feature extraction and unsupervised route planning for Maritime Autonomous Surface Ships (MASS). The method uses Automatic and Adaptive Dynamic Time Warping (AADTW), Spectral Clustering with Affinity Feature (SCAF), and a route optimization algorithm based on dynamic programming to extract features, obtain movement patterns, and plan routes. The proposed method outperforms existing methods by considering the impact of hidden factors and providing different routes for different types of MASS. 
The work in~\cite{li2023ais} systematically analyzes the performance of twelve ship trajectory prediction methods, including classical machine learning and emerging deep learning techniques. It compares twelve methods across three AIS datasets, representing different maritime traffic scenarios, and evaluates their effectiveness based on six indexes. The study concludes that traditional machine learning-based trajectory prediction methods struggle to meet the rising demands for accuracy and real-time performance, leading to increased interest in and promising results from deep learning-based approaches.

A maritime traffic route extraction approach based on multi-dimensional density-based spatial clustering of applications with noise (MD-DBSCAN) was developed in~\cite{huang2023maritime}. The approach incorporates trajectory compression, similarity measures, and extraction of ship trajectory clusters. The approach demonstrates effectiveness in noise reduction and route extraction. The authors in~\cite{wang2021ship} proposed a trajectory clustering method based on Hierarchical Density-Based Spatial Clustering of Applications with Noise (HDBSCAN) and Hausdorff distance to generate a similarity matrix. The method adapts to shape characteristics and exhibits good clustering scalability and improved clustering results compared to DBSCAN, k-means, and spectral clustering algorithms.

Eljabu et al. proposed spatial clustering methods (SPTCLUST and SPTCLUST-II) in~\cite{eljabu2022spatial_I} and~\cite{eljabu2022spatial_II} respectively, for maritime traffic routes extraction from AIS data. The approach consists of data preprocessing, pathfinding, and route extraction without using traditional clustering algorithms. It achieved high F1-scores, 97\% and 99\%, for tankers and cargo maritime traffic routes. 

The study in~\cite{han2021modeling} enhanced the DBSCAN method by integrating the Mahalanobis distance metric for vessel behavior modeling. The proposed methodology includes clustering historical AIS data and detecting anomalies. The study showcases applicability to diverse water regions, contributing to situational awareness, collision prevention, and route planning.

Farahnakian et al.~\cite{farahnakian2023comprehensive} conducted a comprehensive examination of clustering-based techniques, including k-means, DBSCAN, Affinity Propagation (AP), and the Gaussian Mixtures Model (GMM), for detecting abnormal vessel behaviors from AIS data. Results indicate that k-means is particularly effective in detecting dark ships and spiral vessel movements, which is crucial for enhancing maritime safety.
Furthermore, the study~\cite{moavinis2023detection} proposed two methods for trajectory outlier detection, with the first utilizing DBSCAN clustering and Hausdorff distance, and the second employing Support Vector Machine (SVM) classifier and the Generalized Sequence Pattern algorithm. Both models outperform the baselines, with the SVM approach demonstrating superior performance in the identification of traffic patterns and outliers.

The research paper~\cite{liu2021visualization} offers a detailed survey of visual analytics for vessel trajectory data. The authors discuss a variety of methods, including map-based visualization, timeline-based visualization, and interactive visualization. 

The paper~\cite{zhang2022vessel} comprehensively reviews various approaches for vessel trajectory predicting, including clustering algorithms and machine learning algorithms. It also discusses the challenges and future research directions, such as the uncertainty in the data, the dynamic environment, and the computational complexity.

Among the identified challenges, which are subjects of ongoing research and require additional attention, three are worthy of specific mention:
navigating dynamic maritime environments poses a substantial challenge in accurately identifying vessel paths (I); ensuring stability, explainability, and managing the computational cost of the model add further complexity (II); finally, addressing the need for flexibility, scalability, and practical applicability is crucial for a comprehensive solution in the field of vessel path identification (III).
Motivated by these challenges, we aim to develop a framework that focuses on vessel path identification and potentially tackling such challenges faced in maritime transportation.

\section{Methodology}\label{Sec:Methodology}
\begin{figure}[htb]
\centering
    \begin{subfigure}[b]{0.49\linewidth}
        \centering
        \includegraphics[width=0.79\linewidth]{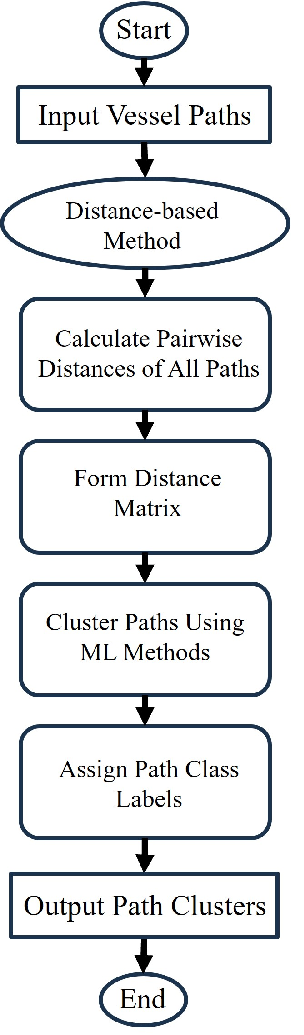}
        \caption{}
        \label{fig:fw_distance}
    \end{subfigure}
\hfill
\centering
     \begin{subfigure}[b]{0.49\linewidth}
        \centering
        \includegraphics[width=0.77\linewidth]{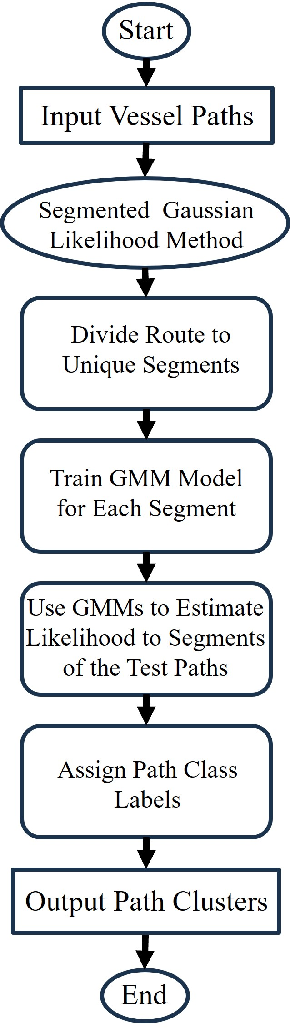}
        \caption{}
        \label{fig:fw_likelihood}
    \end{subfigure}
\caption{Framework of vessel path identification. (a) Flowchart of distance-based method. (b) Flowchart of segmented Gaussian likelihood method.}
\label{fig:framework_path_id}
\end{figure}

This section covers the theoretical background and description of our proposed framework's underlying methodology.
The framework of vessel path identification is depicted in Figure~\ref{fig:framework_path_id}.

\subsubsection{Problem Formulation}\label{Sec:prob_form}
The equations~(\ref{eq:Path_Clust1}-\ref{eq:Path_Clust3}) serve as a mathematical representation to describe the clustering of vessel paths.\\
It is worth mentioning that the clustering process is conducted sequentially, point by point, while the labeling of path classes is performed for the entire voyage. Therefore, each voyage has a single label of path class.

\begin{equation} \label{eq:Path_Clust1}
Voyages \in Path\ Classes\\
\end{equation} 
\begin{equation} \label{eq:Path_Clust2}
Voyages=\{ts_1[p_1,\dots, p_n],\dots,ts_j[p_1,\dots,p_n] \} \\
\end{equation} 
\begin{equation} \label{eq:Path_Clust3}
Path\ Class\ Set = \{class_1,\dots, class_k\}
\end{equation}
where:\\
$Voyages$: a collection of time series, each representing a voyage of the vessel taken following a given path, with a predicted clustered class. \\
\textit{$ts_j$}: a time series corresponding to voyage $j$, i.e., a sequence of $n$ data points, where each data point $p$ represents the vessel position and is defined by a pair of coordinates, namely latitude and longitude.\\
$n$: the number of time steps (duration) of each voyage, which can differ from one voyage to another.\\
$j$: a total number of voyages.\\
$Path\ Class\ Set$: a set of $k$ classes into which the path of voyages are being clustered.

\subsubsection{Distance-Based Method}\label{Sec:Distance_method}
The similarity between two paths is measured by the average nearest neighbor distance (ANND), as shown in Eq.~(\ref{eq:nearest_distance}).\\
\begin{equation} \label{eq:nearest_distance}
ANND (i, j) = \frac{1}{n_i} \sum_{k=1}^{n_i} Distance(P_i^k,  NN(P_j^k))
\end{equation}
where:\\
$ANND(i, j)$: is the average nearest neighbor distance between path $i$ and path $j$, present in the distance matrix at row $i$ and column $j$. It is a symmetric, meaning that $ANND(i,j)$ is the same as $ANND(j,i)$\\
$Distance(P_i^k, NN(P_j^k))$ : The distance between the $k^{th}$ point in path $i$, denoted as $P_i^k$, and its corresponding nearest neighbor point in path $j$, indicated as $NN(P_j^k)$. $n_{i}$ is the total number of points in path $i$.\\ 
The measure $Distance$ is an Euclidean distance. However, for longer curved routes, Haversine or Great-circle distance would be more suitable.

\begin{figure}[htb]
  \centering
  \includegraphics[width=\linewidth]{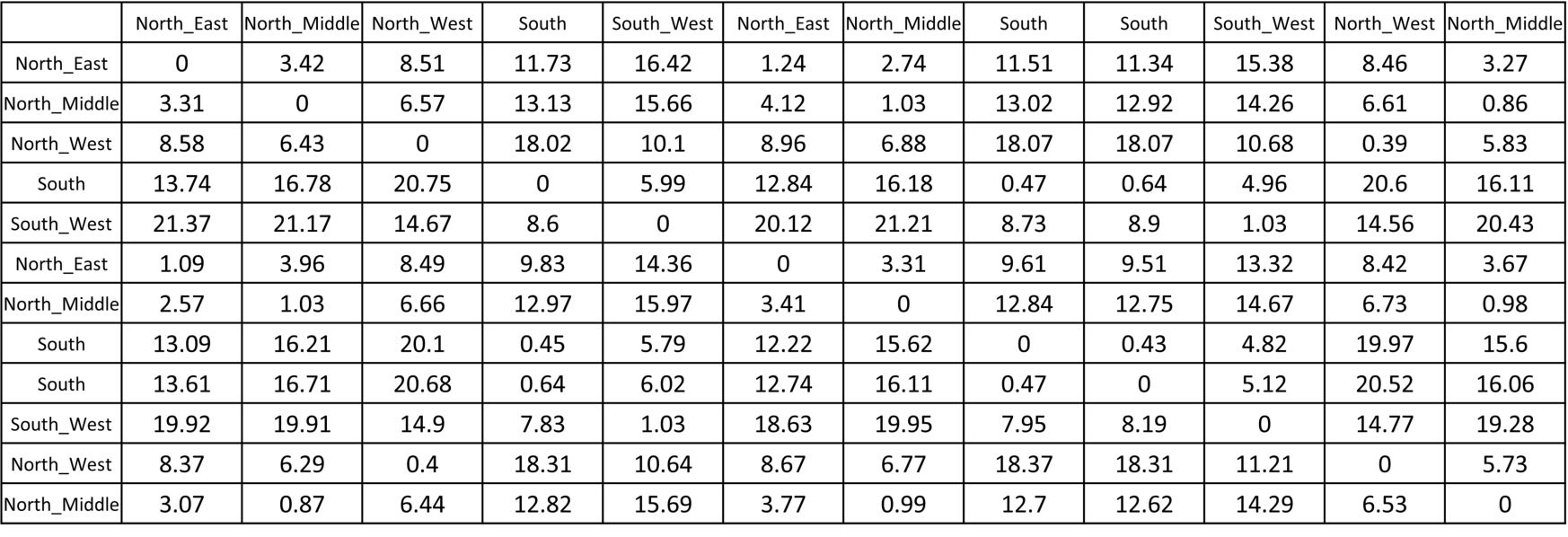}
  \caption{Part of the distance matrix, in this case showing 12 paths.}
  \label{fig:distance_matrix}
\end{figure}

The ANND, as expressed in Eq.~(\ref{eq:nearest_distance}), is computed by averaging the distances between each point in one path and its nearest neighbor in the other path.\\
Then, the similarity value (i.e., ANND) of this pair of paths is stored as an element in the distance matrix.\\
A lower ANND indicates that the paths within a cluster are more similar.
The distance matrix will have dimensions $(m \times m)$ , where $m$ is the number of paths.\\
For instance, the computed distance matrix for a set of 12 paths is illustrated in Figure~\ref{fig:distance_matrix}.

After the construction of the distance matrix, the machine learning (ML) technique is applied to cluster the paths based on their corresponding values in this distance matrix. The ML techniques that we used are k-means, Gaussian Mixture Model (GMM), and hierarchical clustering. \\

\subsubsection{Segmented Gaussian Likelihood Method}\label{Sec:Gauss_method}
In addition to identifying the vessel's path, to better understand how the vessel changes its paths, we employ Gaussian distributions on several distinct segments of the vessel route.
This technique can be summarized as follows:
\begin{itemize}
    \item Utilize a training dataset comprising vessel position information that should adequately represent all potential paths of the vessel route.
    \item Divide the route into different distinct segments. 
    \item Train a single GMM model for each segment to find the Gaussian distributions of all segments of the route. 
    \item Estimate likelihoods of the segments by using the trained GMM models with their corresponding segments of each vessel voyage in the test dataset. 
    \item Label the path classes based on the estimated likelihood at the unique segments of the route.
\end{itemize}

\section{Case Study}\label{Sec:Case_study}

In this section, we describe the case study, including the data collection, preprocessing, and analysis. 

\subsection{Data Collection} 
In this study, we utilized a dataset that was collected from a ship named Cinderella II, which is a passenger ship that operates in the Stockholm archipelago.
The dataset spans over five months, from July to November 2022.
The dataset comprises information on 124 voyages of this vessel, connecting the two main ports of Vaxholm in the east and Sodra in the west.

Additional information about Cinderella II can be found in~\cite{marinetrafffic_Cind}. For instance, the specifications of this vessel are 19 meters in length and 6.41 meters in breadth, boasting a gross tonnage of 68. Operating under the flag of Sweden, its maritime mobile service identity (MMSI) is 265513810 with a reported draught of 2.5 meters and a recorded average speed of 8.2 knots (4.2 m/s).

\begin{figure}[htb]
  \centering
  \includegraphics[width=0.6\linewidth]{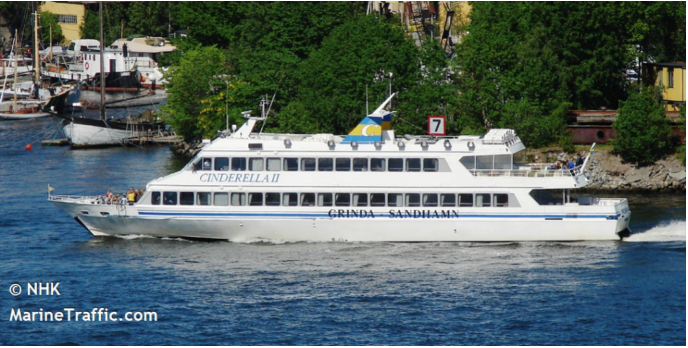}
  \caption{Image of the Cinderella II ship~\cite{marinetrafffic_Cind}.}
  \label{fig:ship_image}
\end{figure}

\begin{figure}[htb]
  \centering
  \includegraphics[width=0.6\linewidth]{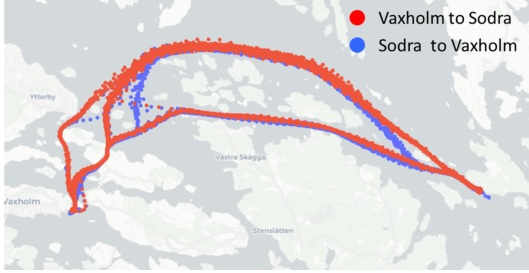}
  \caption{Map of vessel route.}
  \label{fig:routes_map}
\end{figure}

\begin{figure}[htb]
\centering
\begin{subfigure}[b]{0.48\linewidth}
\centering
    \includegraphics[width=\linewidth]{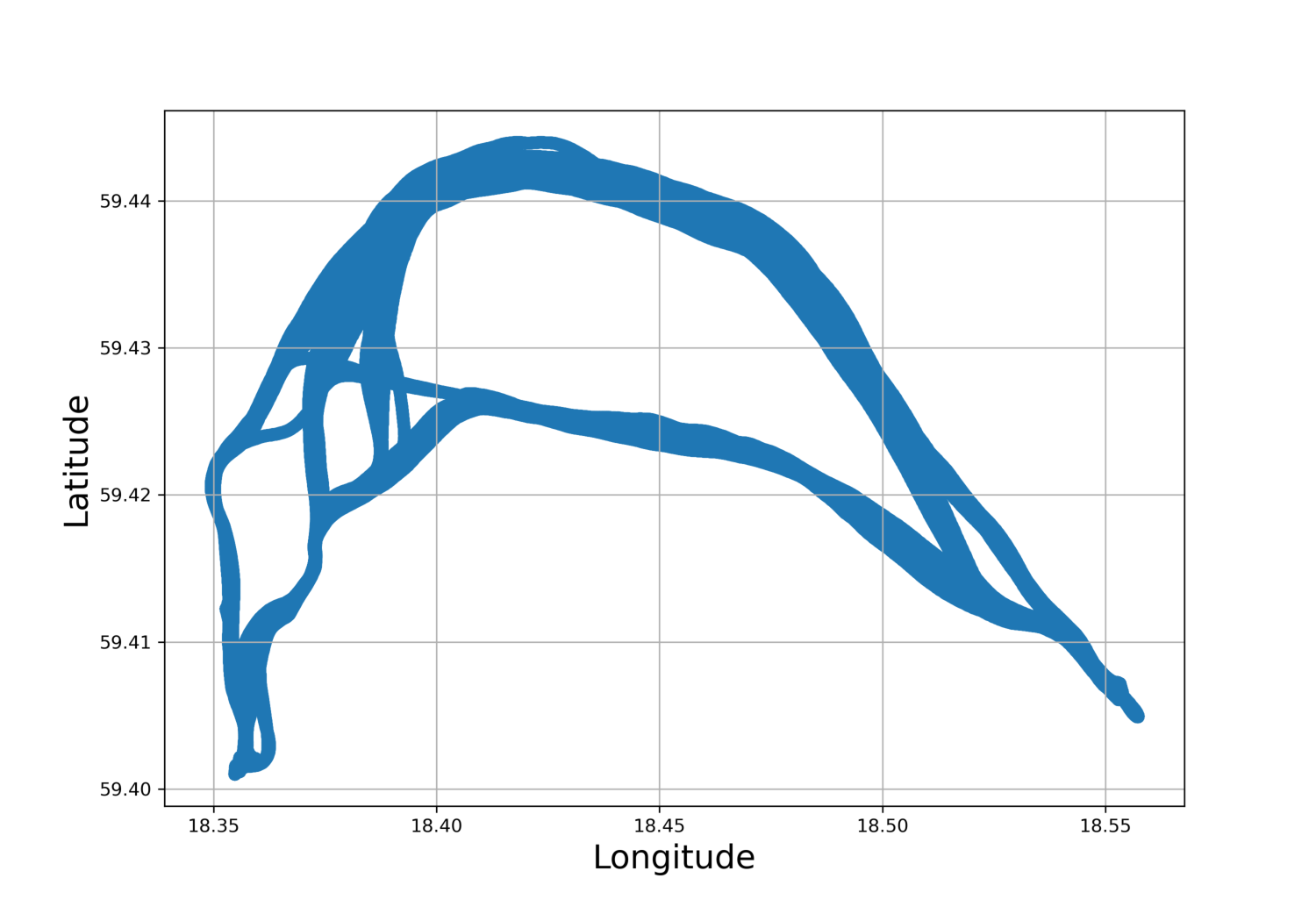}
    \caption{The vessel route.}
    \label{fig:ship_route}
\end{subfigure}
\hfill
\centering
\begin{subfigure}[b]{0.48\linewidth}
\centering
    \includegraphics[width=\linewidth]{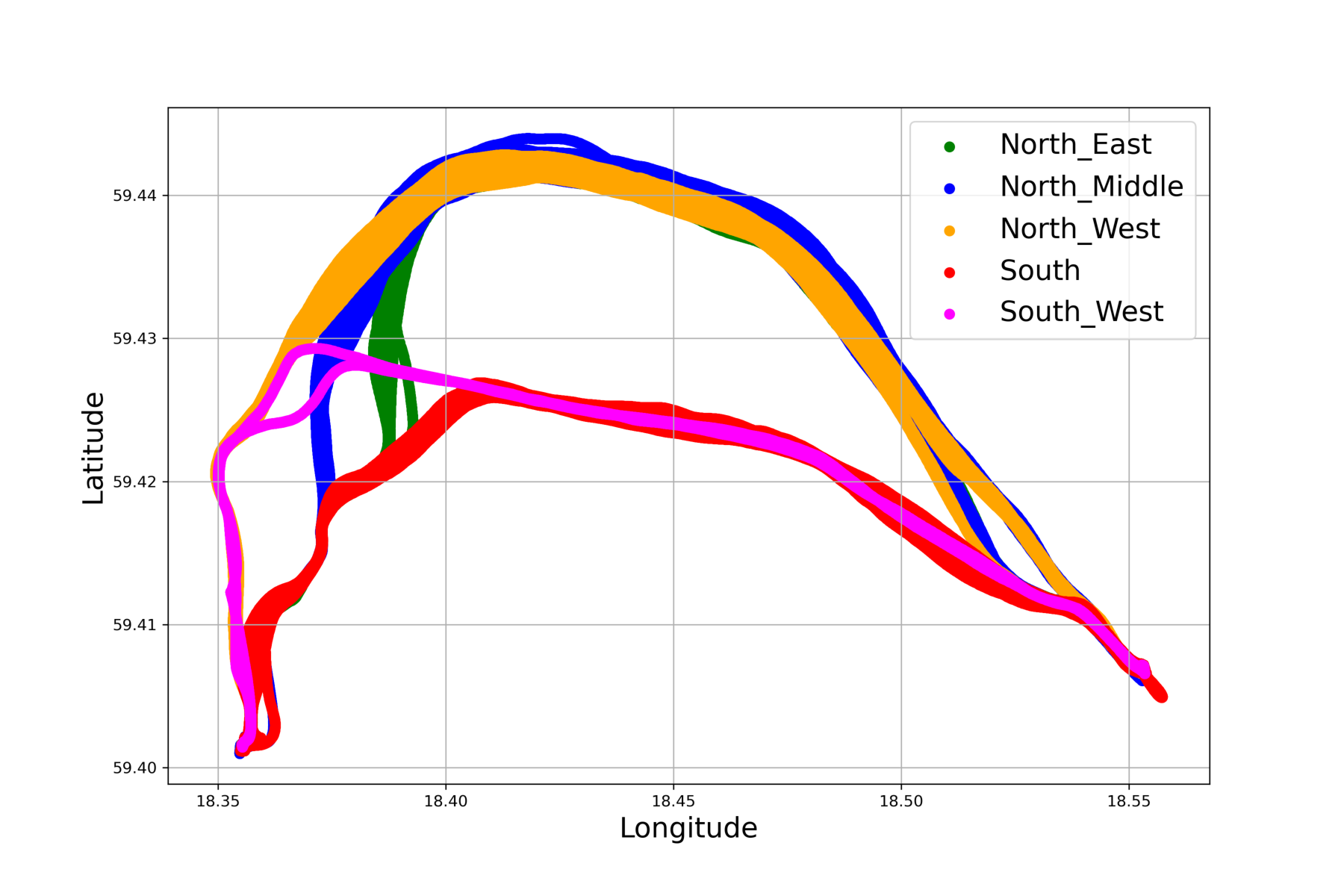}
    \caption{Main classes of path}
    \label{fig:1plt_path5cls}
\end{subfigure}
\caption{The vessel route before and after applying path identification}
\label{Cind_route}
\end{figure}

\subsection{Data Preparation and Analysis}\label{data_prep}
\begin{figure}[htb]
    \centering
    \begin{subfigure}{1.1\linewidth}
        \centering
        \includegraphics[width=\linewidth]{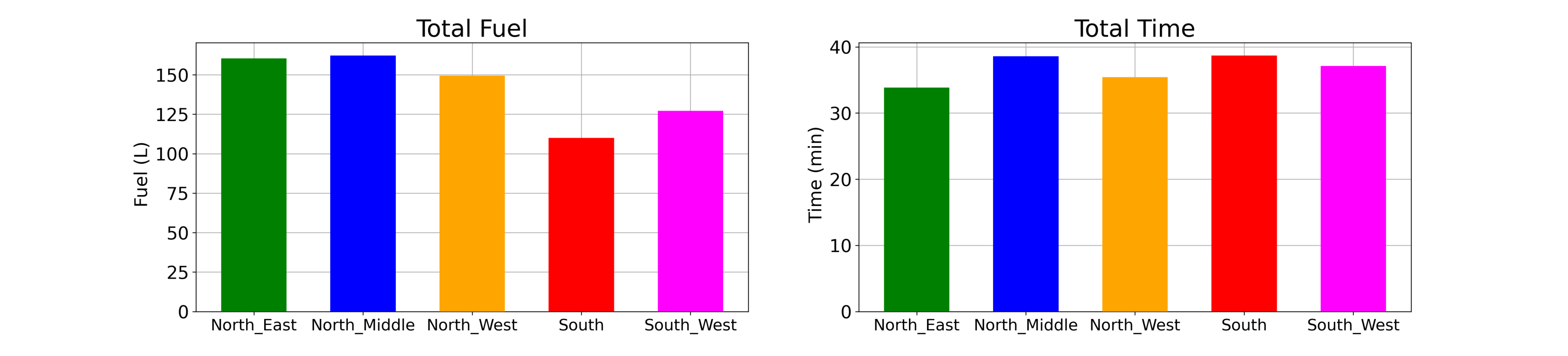}
    \end{subfigure}
    
    \begin{subfigure}{1.1\linewidth}
        \centering
        \includegraphics[width=\linewidth]{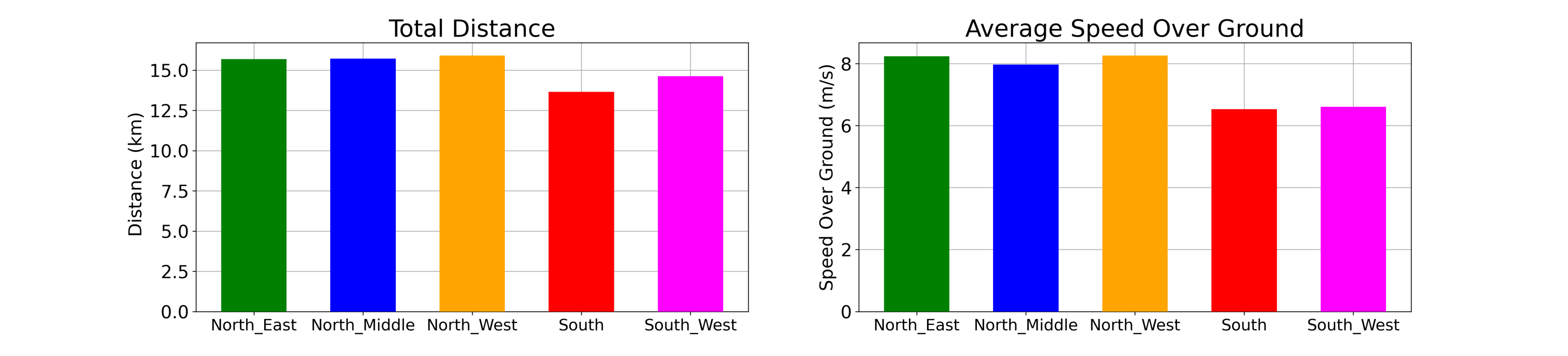}
        \label{fig:distance_speed_path5cls}
    \end{subfigure}
\caption{Average fuel and time, distance, and speed of five path classes}
\label{fig:FTDS_figure}
\end{figure}

In our approach, we emphasize the significance of data representation.
As a result, we group the path points based on their timestamps with a resolution of one second and store these grouped path points with distinctive Voyage IDs. Afterward, these paths are ready to be processed by the path clustering approach to determine the overall path class.
In order to exploit the resulting class information, statistical analysis is conducted to determine differences in vessel paths concerning fuel, time, distance, and speed.
The histograms representing these quantities across various path classes can be found in Figure~\ref{fig:FTDS_figure}. Notably, when the vessel traverses the shorter southern paths, it employs slower speeds, effectively reducing fuel consumption without significantly impacting travel time.

\begin{figure}[htb]
  \centering
  \includegraphics[width=\linewidth]{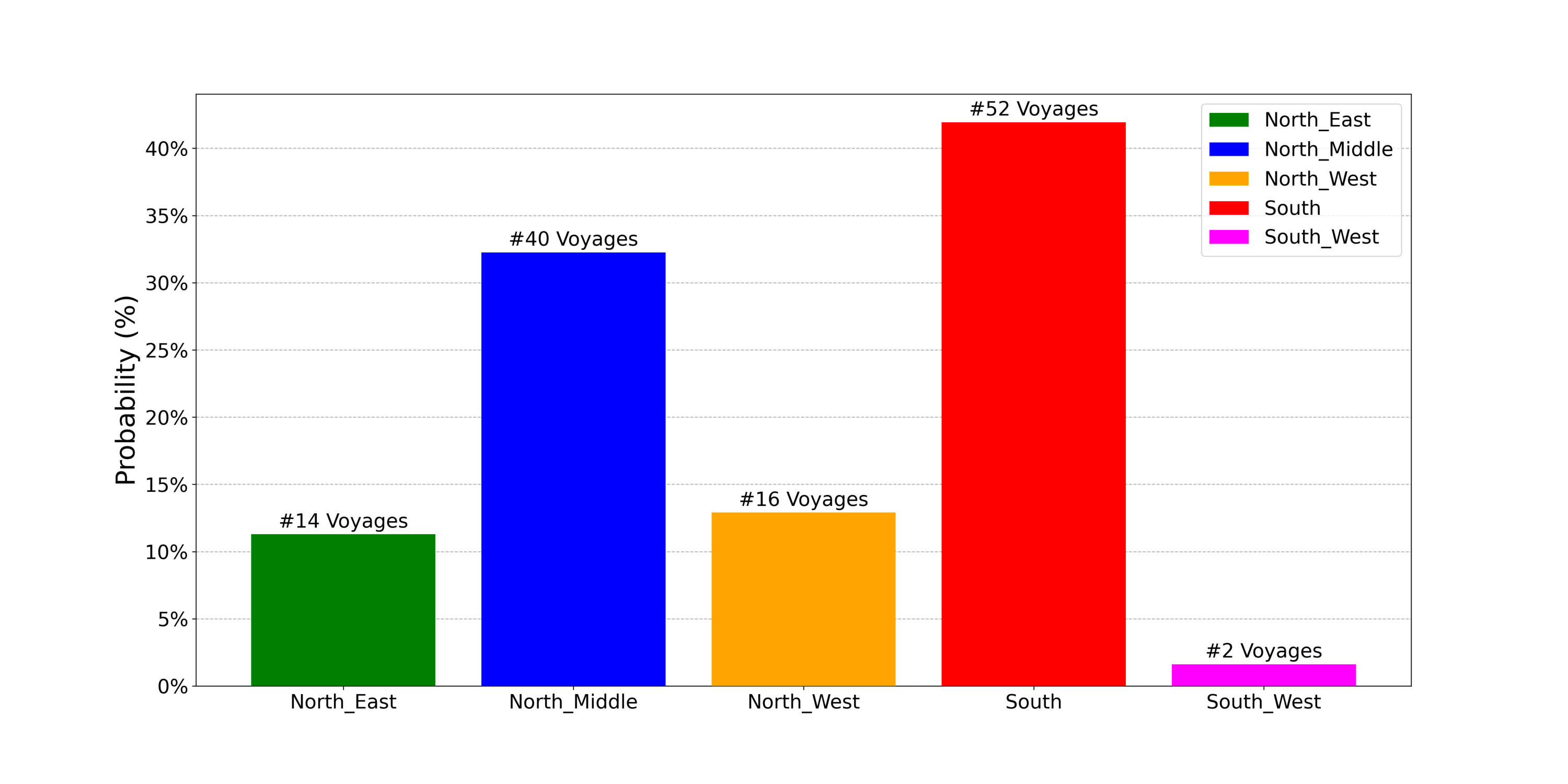}
  \caption{Distribution of voyages across the five path classes}
  \label{fig:stats_path5cls}
\end{figure}

\section{Results and Discussion}\label{Sec:Results}

In this section, we present the results of our spatial clustering approach for vessel path identification and discuss the implications of these findings. The approach involved the utilization of position information and various clustering techniques, specifically k-means, hierarchical clustering, and Gaussian distributions clustering, with the dataset containing 124 voyages.

\subsection{Evaluation of Path Clustering}\label{eval_metrics}

The results of vessel path identification are evaluated through visual inspection and tabulation using metrics such as confusion matrix, precision, recall, and F1-score~\cite{yan2021artificial}. 

The hits and messes of path clustering are presented by the confusion matrix.
For our results of path clustering, the confusion matrix is a one-vs-one type matrix.
Then, the confusion matrix is converted into a one-vs-all type matrix (binary-class confusion matrix)~as shown in Eq.~(\ref{eq:cofus_mat_2class}), for calculating class-wise metrics like precision, recall, and F1-score. 

\begin{equation} \label{eq:cofus_mat_2class}
\begin{array}{c|cc}
& \text{Pred. Pos.} & \text{Pred. Neg.} \\
\hline
\text{Act. Pos.} & \text{TP } & \text{FN } \\
\text{Act. Neg.} & \text{FP } & \text{TN }
\end{array}
\end{equation}
where True Positives (TP), False Positives (FP), True Negatives (TN), and False Negatives (FN) are determined by comparing the predicted (Pred.) and actual (Act.) path classes.

The following performance metrics were used:
\begin{itemize}
    \item Precision: the ratio of true positives to the total number of predicted positives.
    \item Recall: the ratio of true positives to the total number of actual positives in the data.
    \item F1-score: the harmonic mean of precision and recall. 
\end{itemize}

The equations for precision, recall, and F1-score are shown in Eqs.~(\ref{eq:Precision}), (\ref{eq:Recall}), and (\ref{eq:F1score}).

\begin{equation}\label{eq:Precision}
Precision = \frac{TP}{TP + FP}
\end{equation} 

\begin{equation}\label{eq:Recall}
Recall = \frac{TP}{TP + FN}
\end{equation}

\begin{equation}\label{eq:F1score}
F1\text{-}score= 2\times\frac{(Precision \times Recall)}{(Precision + Recall)}
\end{equation}

\subsection{Discussion}\label{discussion}

Table~\ref{Table:Result_Kmean_GMM_Summary} shows the results of applying k-means or Gaussian Mixture Model (GMM) models to identify the vessel paths from the distance matrix in the distance-based method of the path identification approach.
Notably, the paths with classes of North-West, South, and South-West achieved an F1-score of 1.0, indicating that the approach correctly identified all the paths of these classes.

\begin{figure}[htb]
  \centering
  \includegraphics[width=\linewidth]{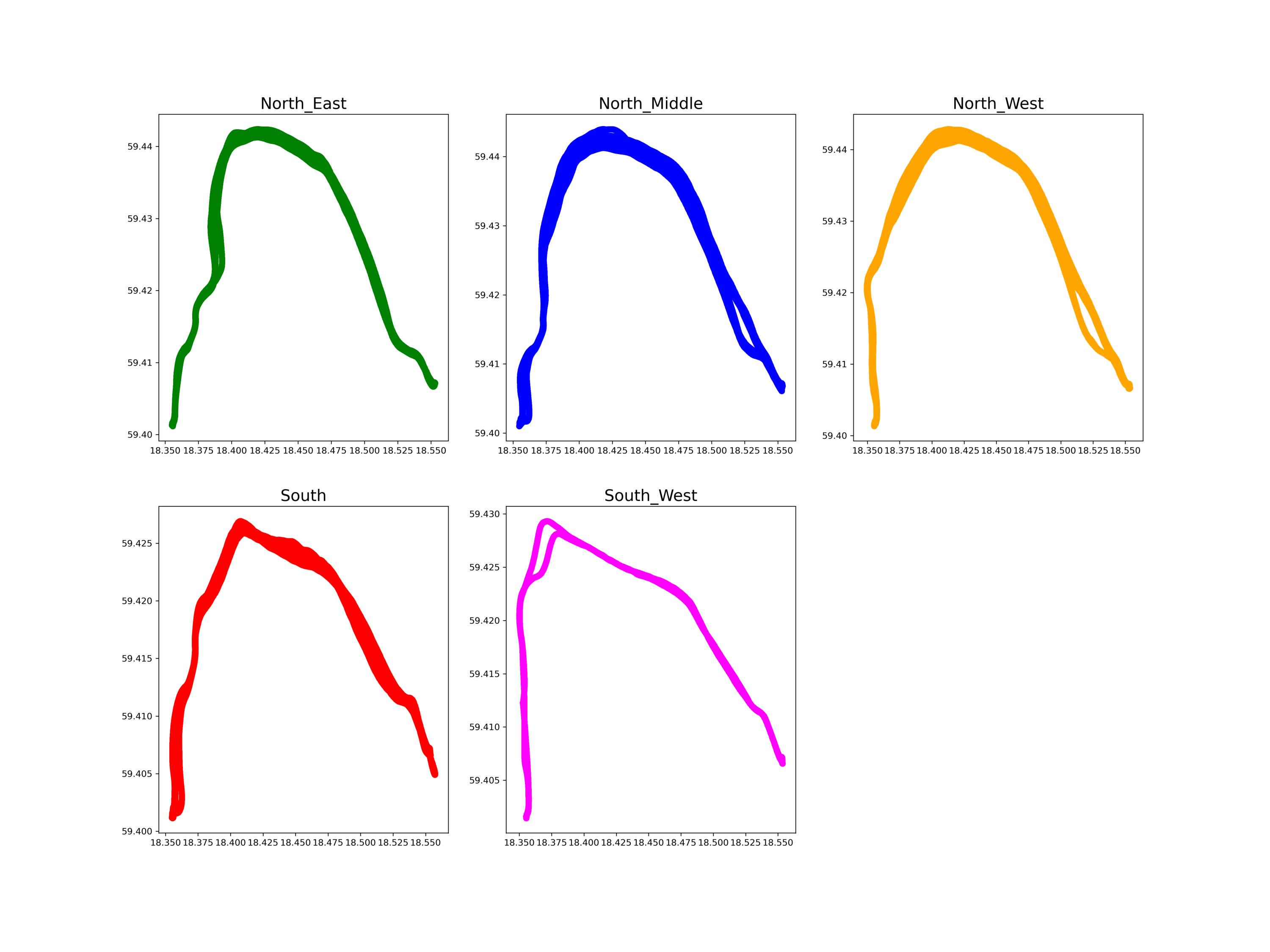}
  \caption{Display of the five path classes (ground truth).}
  \label{fig:5plts_path5cls}
\end{figure}

\begin{figure}[htb]
  \centering
    \includegraphics[width=1.1\linewidth]{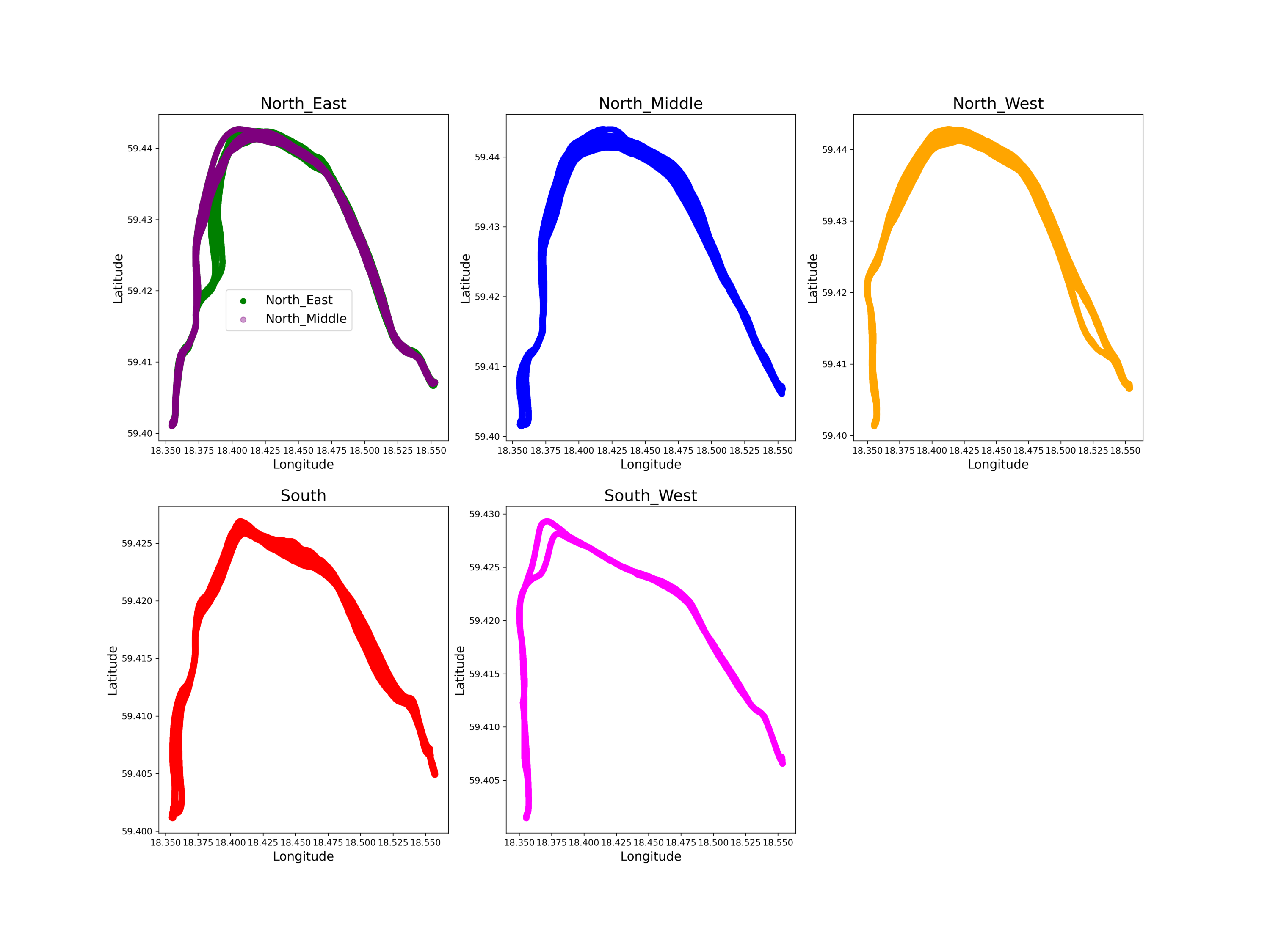}
  \caption{Results of both k-means and GMM clustering to five path classes.}
  \label{fig:Kmean_GMM_clus_path5cls}
\end{figure}

\begin{figure}[htb]
  \centering
  \includegraphics[width=1.1\linewidth]{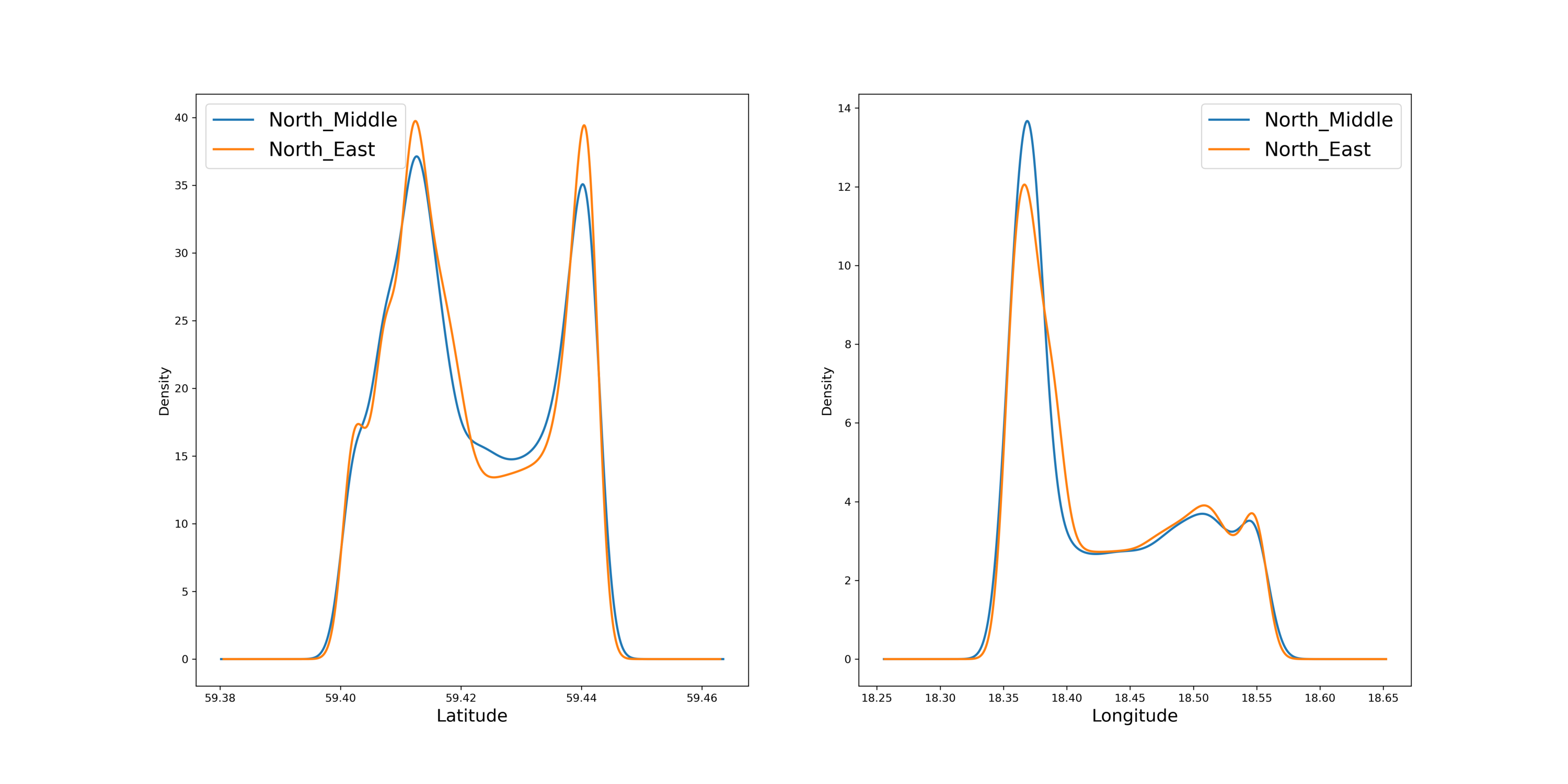}
  \caption{Probability distribution of location coordinates for mis-clustered paths by both k-means and GMM.}
  \label{fig:Fig_Cind_5paths_PDF_Misclustered}
\end{figure}

In contrast, the North-East and North-Middle paths exhibited lower F1 scores compared to other classes. 
The path class of North-Middle is the most challenging path to identify since six such paths have been clustered as North-East, as can be seen by comparing Figures~\ref{fig:5plts_path5cls} and~\ref{fig:Kmean_GMM_clus_path5cls}, which are the visualization for all the paths, color-marked based on their ground truth classes.
Figure~\ref{fig:Fig_Cind_5paths_PDF_Misclustered} illustrates the probability distribution of mis-clustered paths with respect to latitude and longitude coordinates. It is obvious that these paths have nearly identical coordinates, which makes them challenging paths to cluster with k-means or GMM.
This suggests that there is still room for improvement by using other ML clustering methods.

\begin{figure}[htb]
  \centering
  \includegraphics[width=1.1\linewidth]{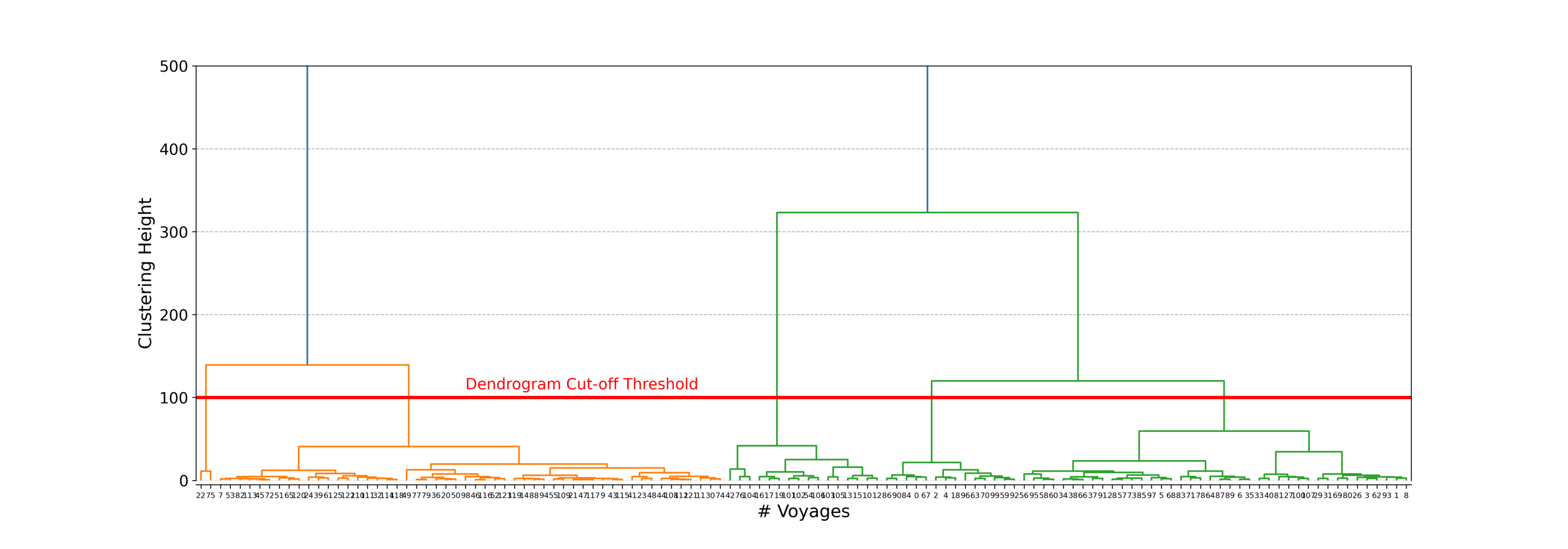}
  \caption{Results of hierarchical clustering to five path classes.}
  \label{fig:Hierch_clus_path5cls}
\end{figure}

Table~\ref{Table:Result_Hierch_Summary} presents the results of employing hierarchical clustering to the distance matrix in the distance-based method for clustering the vessel paths. In hierarchical clustering, there is a parameter called "Dendrogram Cut-off threshold," and its value should be selected depending on the number of path classes. Hence, as illustrated in Figure~\ref{fig:Hierch_clus_path5cls}, this parameter is denoted by the Y-axis as a clustering height, and its value is set to 100 for clustering the vessel path into five classes.

\begin{figure}[htb]
  \centering
  \includegraphics[width=\linewidth]{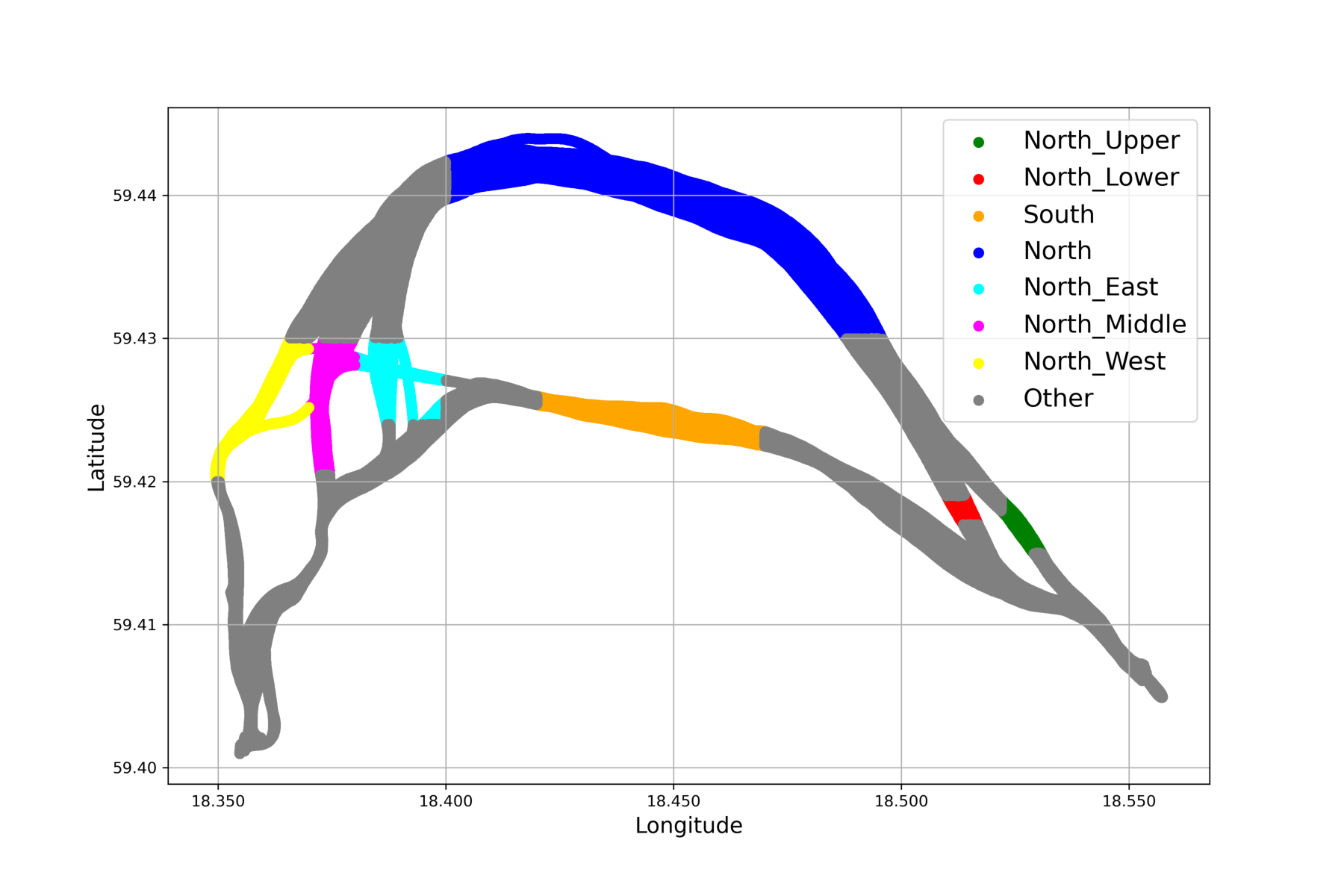}
  \caption{Distribution of the route into eight segments.}
  \label{fig:Distribution_path8cls}
\end{figure}

\begin{figure}[htb]
  \centering
  \includegraphics[width=\linewidth]{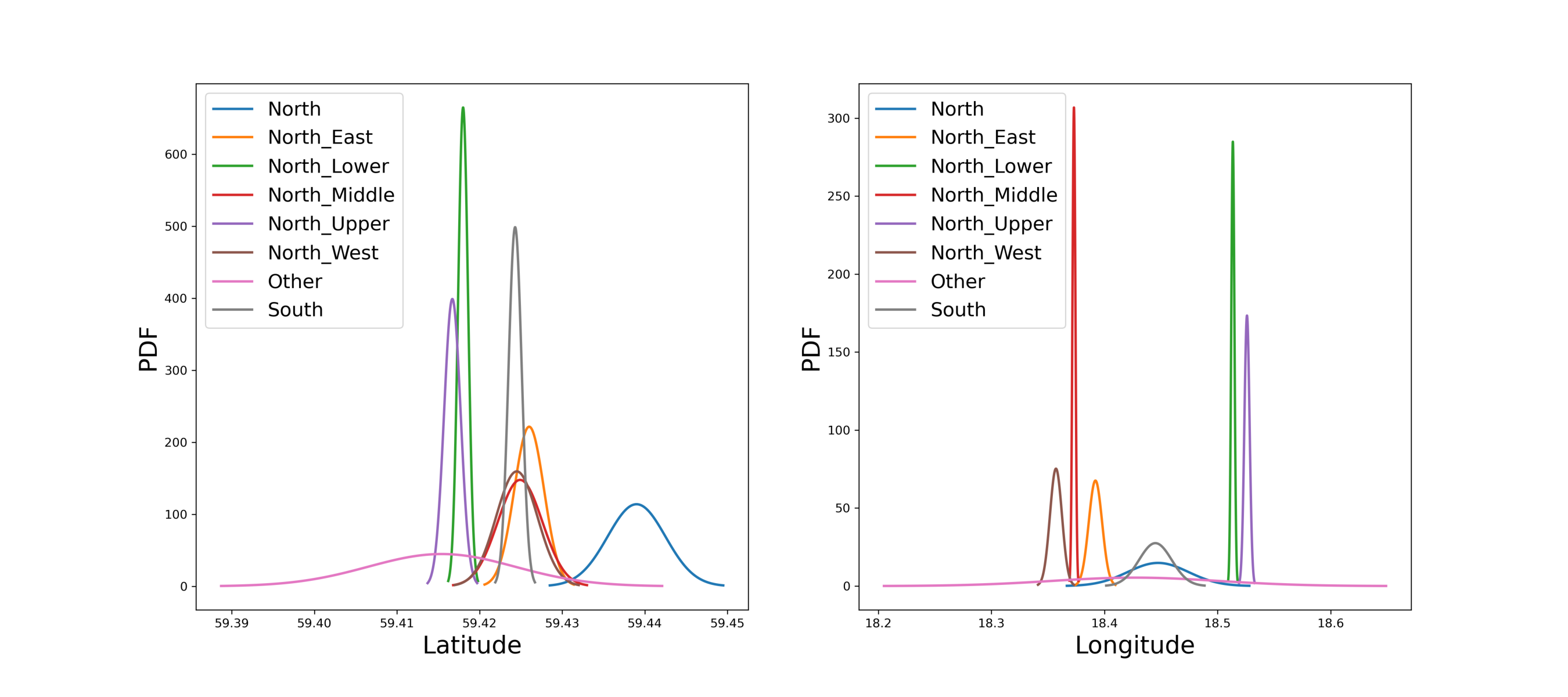}
  \caption{Probability distributions of location coordinates for the eight segments of the route.}
  \label{fig:Prob_path8cls_latlon}
\end{figure}

\begin{figure}[htb]
  \centering
  \includegraphics[width=\linewidth]{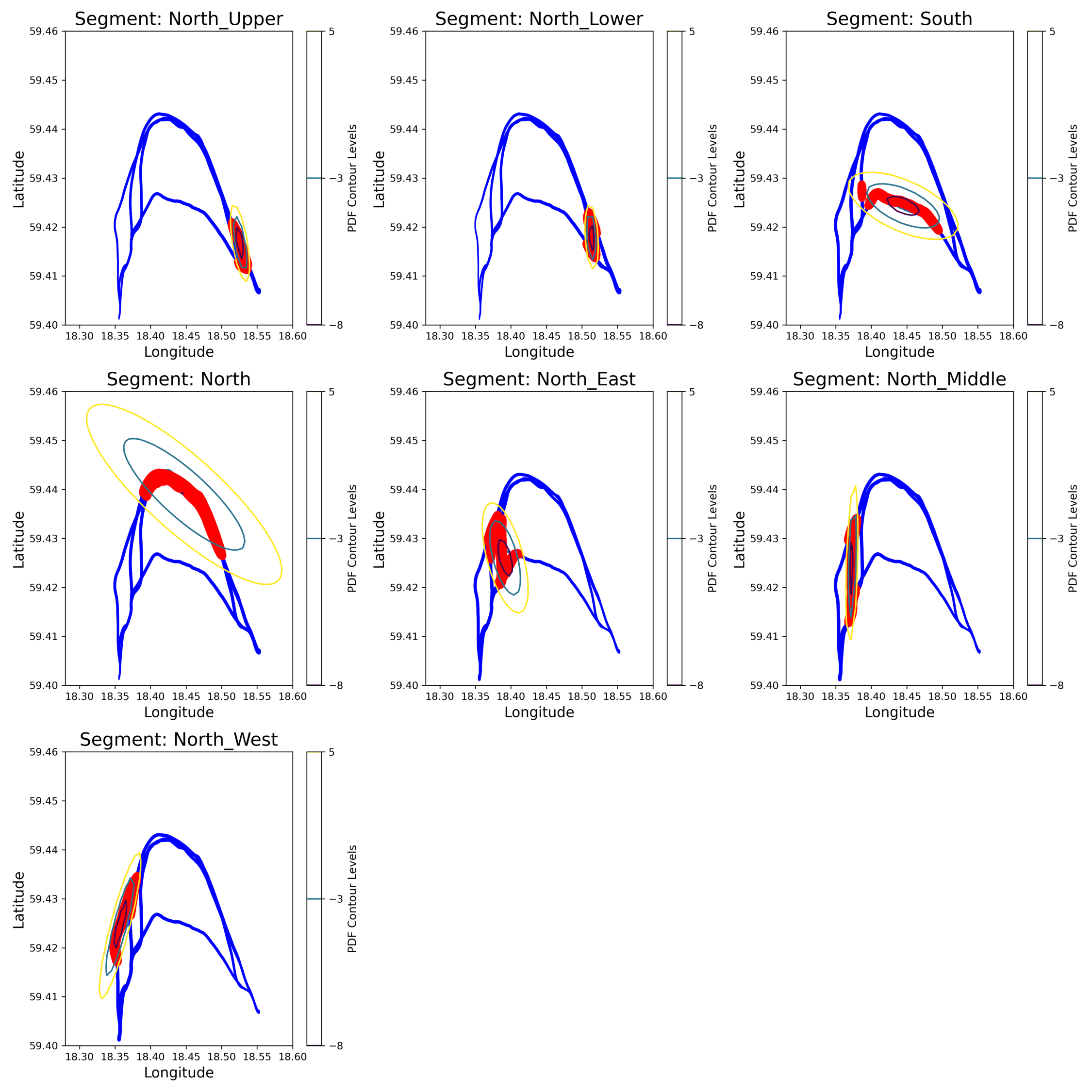}
  \caption{Gaussian distributions for seven segments of the route.}
  \label{fig:Gauss_path8cls}
\end{figure}

Remarkably, all path classes achieved an F1-score of 1, indicating that hierarchical clustering successfully identified all paths with high accuracy from the distance matrix using the distance-based method.
This suggests that the choice of ML clustering technique with the distance matrix can influence the accuracy of path identification.
Table~\ref{Table:Result_Gauss_Summary} displays the outcomes of clustering, now by applying the segmented likelihood Gaussian method. This method achieved perfect precision, recall, and F1-score for all path classes.\\
Figures~\ref{fig:Distribution_path8cls}, \ref{fig:Prob_path8cls_latlon}, and \ref{fig:Gauss_path8cls} present visualizations for the segmented Gaussian likelihood method.

The accuracy in results by hierarchical and segmented Gaussian likelihood clustering for path classes indicates the efficacy of the developed approach of spatial clustering for vessel path identification.
\begin{table}[htb]
    \centering
    \caption{Result of implementing of both k-means and GMM clustering to five path classes}
    \label{Table:Result_Kmean_GMM_Summary}
    
    \begin{subtable}{\linewidth}
        \centering
        \caption{Precision, Recall, and F1-score}
        \begin{tabular}{cccc}
            \toprule
            \textbf{Paths} & \textbf{Precision} & \textbf{Recall} & \textbf{F1-score} \\
            \midrule
            North-East (NE)    & 0.7 & 1   & 0.824 \\
            North-Middle (NM)  & 1   & 0.85 & 0.919 \\
            North-West (NW)   & 1   & 1   & 1     \\
            South (S)      & 1   & 1   & 1     \\
            South-West (SW)   & 1   & 1   & 1     \\
            \bottomrule
        \end{tabular}
    \end{subtable}
    
    \vspace{1em} 
    
    \begin{subtable}{\linewidth}
        \centering
        \caption{Confusion Matrix}
        \begin{tabular}{ccccccc}
            \toprule
            \multirow{2}{*}{\textbf{Actual}} & \multicolumn{5}{c}{\textbf{Predicted}} & \multirow{2}{*}{\textbf{Total}} \\
            \cline{2-6}
            & \textbf{NE} & \textbf{NM} & \textbf{NW} & \textbf{S} & \textbf{SW} & \\
            \midrule
            NE & 14 & 0 & 0 & 0 & 0 & 14 \\
            NM & 6  & 34 & 0 & 0 & 0 & 40 \\
            NW & 0  & 0  & 16 & 0 & 0 & 16 \\
            S  & 0  & 0  & 0 & 52 & 0 & 52 \\
            SW & 0  & 0  & 0 & 0 & 2 & 2 \\
            \textbf{Total} & 20 & 34 & 16 & 52 & 2 & 124 \\
            \bottomrule
        \end{tabular}
    \end{subtable}
\end{table}


\begin{table}[htb]
    \centering
    \caption{Result of implementing both hierarchical clustering and segmented likelihood Gaussian clustering to five path classes}
    \label{Table:Result_Hierch_Summary}
    \label{Table:Result_Gauss_Summary}
    
    \begin{subtable}{\linewidth}
        \centering
        \caption{Precision, Recall, and F1-score}
        \begin{tabular}{cccc}
            \toprule
            \textbf{Paths} & \textbf{Precision} & \textbf{Recall} & \textbf{F1-score} \\
            \midrule
            North-East (NE)    & 1 & 1   & 1 \\
            North-Middle (NM)  & 1 & 1 & 1 \\
            North-West (NW)   & 1   & 1   & 1     \\
            South (S)      & 1   & 1   & 1     \\
            South-West (SW)   & 1   & 1   & 1     \\
            \bottomrule
        \end{tabular}
    \end{subtable}
    
    \vspace{1em} 
    
    \begin{subtable}{\linewidth}
        \centering
        \caption{Confusion Matrix}
        \begin{tabular}{ccccccc}
            \toprule
            \multirow{2}{*}{\textbf{Actual}} & \multicolumn{5}{c}{\textbf{Predicted}} & \multirow{2}{*}{\textbf{Total}} \\
            \cline{2-6}
            & \textbf{NE} & \textbf{NM} & \textbf{NW} & \textbf{S} & \textbf{SW} & \\
            \midrule
            NE & 14 & 0 & 0 & 0 & 0 & 14 \\
            NM & 0  & 40 & 0 & 0 & 0 & 40 \\
            NW & 0  & 0  & 16 & 0 & 0 & 16 \\
            S  & 0  & 0  & 0 & 52 & 0 & 52 \\
            SW & 0  & 0  & 0 & 0 & 2 & 2 \\
            \textbf{Total} & 14 & 40 & 16 & 52 & 2 & 124 \\
            \bottomrule
        \end{tabular}
    \end{subtable}
\end{table}


\section{Conclusion}\label{Sec:Conclusion}

The proposed approach is able to identify the vessel paths with partially defined or unknown paths.
In the distance-based method, the hierarchical clustering used in the approach outperforms k-means and GMM clustering techniques.

The approach of hierarchical clustering includes a user-customizable parameter, a cut-off threshold, which allows desired control for the number of path classes, enhancing the flexibility and adaptability of the proposed approach.

In the distance-based method, adopting ANND as a measure of similarity makes path clustering less affected by noise or outliers and provides a more intuitive interpretation of path similarity, ultimately enhancing the robustness and interpretability of our approach.

The segmented Gaussian likelihood method is particularly useful for identifying and analyzing the vessel path alterations at different segments of the vessel route.

The proposed approach is computationally efficient and has the potential to be a valuable tool for planning vessel paths. Accurate path identification can contribute to safer and more efficient maritime transportation practices, aiding in route planning, collision avoidance, and navigation optimization.

Nevertheless, the framework has some potential limitations, such as the segmented Gaussian likelihood method exhibiting sensitivity to segment definition, which could affect its salable performance, particularly in complex maritime scenarios.
Moreover, while the study case demonstrates that the framework is computationally efficient, it is essential to discuss any potential scalability issues, especially when dealing with large datasets, since the computational efficiency may vary depending on the dataset size and the nature of the paths.

Further work could explore the scalability and real-world applicability of the proposed clustering approach, as well as its integration with related systems of maritime transportation.

\section*{Acknowledgment}
This research project is funded by Sweden’s innovation agency (Vinnova).\\
The authors wish to thank the diverse group at the Center for Applied Intelligent Systems Research (CAISR), Halmstad University, for helpful discussions.

 
{\appendix[Supplementary Materials]
The source codes that are implemented on Python 3.9.7 to produce the results are available at:~\url{https://github.com/MohamedAbuella/Path_Clustering}.\\
The dataset is private and cannot be shared due to the crucial commercial interests of the startup company operating the iHelm system.}



\bibliographystyle{IEEEtran}
\bibliography{ihelm_bib}

\begin{IEEEbiography}[{\includegraphics[width=1in,height=1.25in,clip,keepaspectratio]{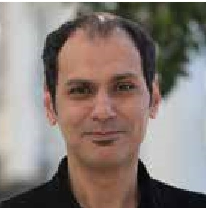}}]
{Mohamed Abuella} received his M.S. and PhD degrees in Electrical and Computer Engineering from Southern Illinois University at Carbondale and University of North Carolina at Charlotte, in 2012 and 2018 respectively. He is a postdoctoral researcher at Halmstad University since 2022.
His research interests include energy analytics and AI for sustainability.
\end{IEEEbiography}

\vspace{-\baselineskip}

\begin{IEEEbiography}[{\includegraphics[width=1in,height=1.25in,clip,keepaspectratio]{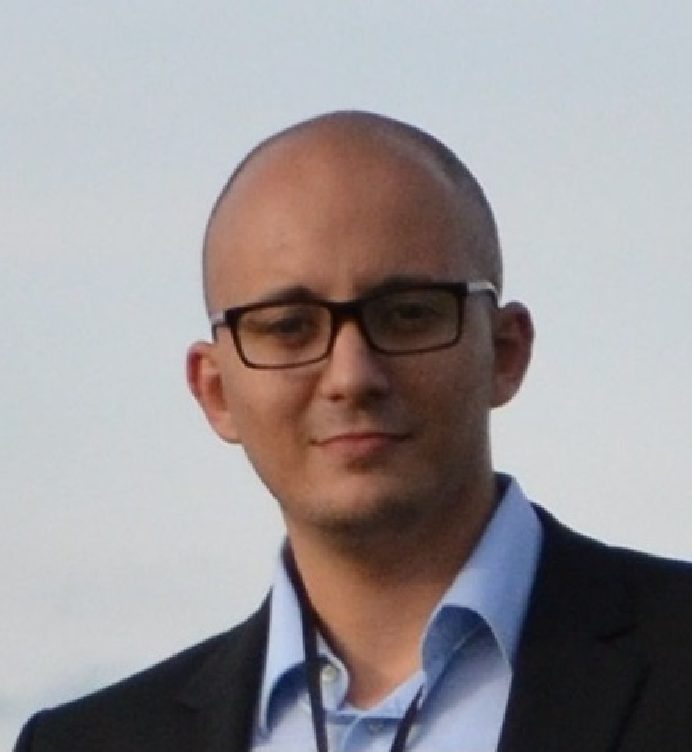}}]
{M. AMINE ATOUI} obtained his Ph.D. degree from LARIS, Polytech' Angers, France, in 2015. Currently, he is affiliated with the Center for Applied Intelligent Systems Research at Halmstad University, Sweden. His research interests encompass probabilistic and explainable Machine Learning, causal and Bayesian Inference, transmission/communication, and automatic control.
\end{IEEEbiography}
\vspace{-\baselineskip}

\begin{IEEEbiography}[{\includegraphics[width=1in,height=1.25in,clip,keepaspectratio]{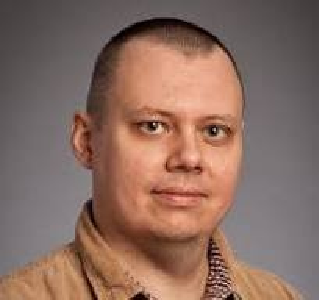}}]
{Slawomir Nowaczyk} is a Professor in Machine Learning at the Center for Applied Intelligent Systems Research, Halmstad University, Sweden. He received his MSc degree from Poznan University of Technology in 2002 and his PhD from the Lund University of Technology in 2008. During the last decades, his research has focused on machine learning, knowledge representation, and self-organising systems. The majority of his work concerns industrial data streams, often with predictive maintenance as the goal. Given that accurate and relevant labels are usually impossible to obtain in such settings, Slawomir’s contributions primarily take advantage of proxy labels, such as transfer learning and multi-task learning, or concern semi-supervised and unsupervised modelling. He is a board member of the Swedish AI Society and a research leader for the School of Information Technology at Halmstad University. Slawomir has led multiple research projects on applying Artificial Intelligence and Machine Learning in different domains, such as transport and automotive, energy, smart cities, and healthcare. In most cases, this research was done in collaboration with industry and public administration organisations – inspired by practical challenges and leading to tangible results and deployed solutions.
\end{IEEEbiography}
\vspace{-\baselineskip}

\begin{IEEEbiography}[{\includegraphics[width=1in,height=1.25in,clip,keepaspectratio]{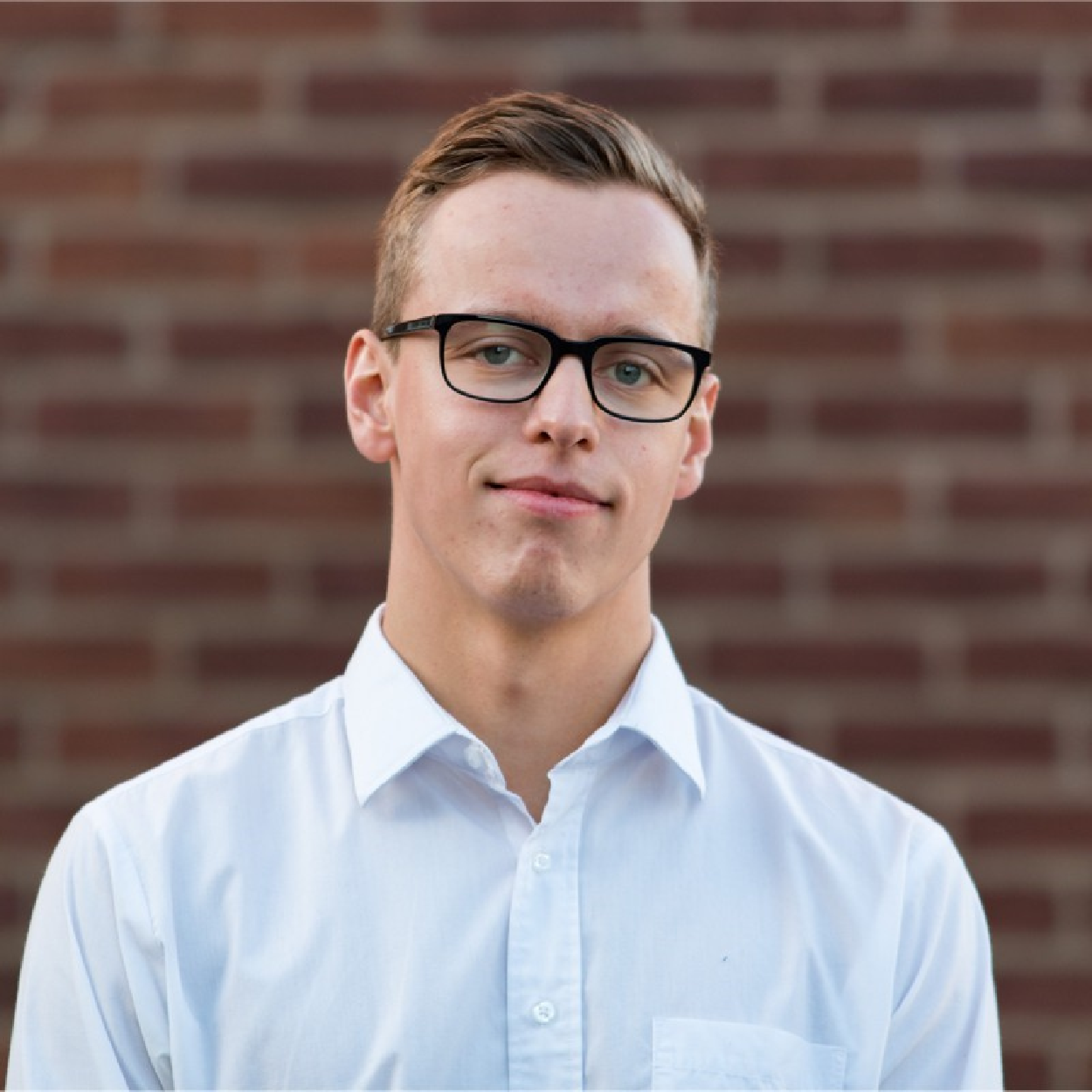}}]
{Simon Johansson} is a MSc graduate of Chalmers University of Technology's program in Engineering Mathematics and Computational Science in 2020, currently works in Cetasol, a marine company specialising in CO2 reduction and energy optimisation for vessels. His practical application of computational methods and dedication to environmental sustainability align with his role, contributing to global efforts to mitigate climate change. Simon's commitment to advancing eco-friendly solutions in the marine industry reflects a seamless integration of academic excellence and real-world impact.
\end{IEEEbiography}
\vspace{-\baselineskip}

\begin{IEEEbiography}[{\includegraphics[width=1in,height=1.25in,clip,keepaspectratio]{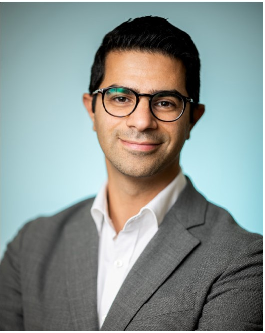}}]
{Ethan Faghani} is the CEO and founder of Cetasol. Before Cetasol, Ethan was Chief Engineer of Automation and AI at Volvo Penta. Ethan has experience working with cutting-edge technologies in other transportation segments in both big enterprises and his own founded startup. Ethan obtained his Ph.D. in mechatronics from UBC and Innovation and Entrepreneurship from Stanford Business School.
\end{IEEEbiography}


 




\vfill

\end{document}